# Tracing the Genealogies of Ideas with Large Language Model Embeddings

Lucian Li, School of Information Science, University of Illinois, Urbana-Champaign

"I have so much to do in unraveling certain human lots, and seeing how they were woven and interwoven, that all the light I can command must be concentrated on this particular web, and not dispersed over that tempting range of relevancies called the universe." – *Middlemarch,* George Eliot

"We shall never disentangle the inextricable web of affinities between the members of any one class; but when we have a distinct object in view, and do not look to some unknown plan of creation, we may hope to make sure but slow progress." - *Origin of Species,* Charles Darwin

In *Darwin's Plots*[1], Gillian Beer examines Darwin's influence on literature as a complex and reciprocal system. Beer identifies in Darwin's writings not only the influence of naturalists and geologists like Lyell, but also the stylistic and lyrical influence of Wordsworth, Coleridge, and Milton. Proceeding onwards, Beer delves into a close reading of how Darwinian metaphors, themes, and worldviews emerge in the works of George Eliot and Thomas Hardy, both correspondents of Darwin who wrote extensive commentaries and reactions to the *Origin of Species*.

As Beer's work shows, there are connections between intellectual figures and avenues for the spread of ideas not possible to observe except through deliberately interdisciplinary efforts. But scholars cannot have expertise in every field and every potential author; experts with training in dozens of subfields and time to read hundreds of thousands of books are in short supply.

Computational methods can enable analysis across some of these boundaries. In this paper, I present a novel method to detect intellectual influence across a large corpus. Taking advantage of the unique affordances of large language models in encoding semantic and structural meaning while remaining robust to paraphrasing, we can search for substantively similar ideas and hints of intellectual influence in a computationally efficient manner. Such a method allows us to operationalize different levels of confidence: we can allow for direct quotation, paraphrase, or speculative similarity while remaining open about the limitations of each threshold.

---

[1] Beer, Gillian. *Darwin's plots: evolutionary narrative in Darwin, George Eliot and nineteenth-century fiction*. Cambridge University Press, 2000.

I apply an ensemble method combining General Text Embeddings (GTE), a state-of-the-art sentence embedding method[2] optimized to capture semantic content and an Abstract Meaning Representation (AMR)[3] graph representation designed to capture structural similarities in argumentation style and the use of metaphor. I apply this method to vectorize sentences from a corpus of roughly 150,000 nonfiction books and academic publications from the 19th century for instances of ideas and arguments appearing in Darwin's publications. This functions as an initial evaluation and proof of concept; the method is not limited to detecting Darwinian ideas but is detecting similarities on a large scale in a wide range of corpora and contexts.

**Related Work:**

Previous attempts to quantify and detect intellectual influence have taken three overall directions: topic modelling, text reuse detection, and word sense similarity.

Studies using topic models generally compare topic distributions across documents or subdocuments. They can capture a zeitgeist of themes and shifting focus but lack granular focus on specific claims. Rockmore et. al.[4] uses topic models to trace the genealogy of national constitutions. In Barron et. al.[5], the authors measure K-L divergence of the Topic Distributions of French Revolutionary speeches. In general, these approaches are generally more effective in a limited context with a controlled set of topics and a high likelihood of influence between documents in the corpus. However, the changes in topic distribution detected by this method may reflect high level shifts in societal context rather than direct influence.

Text reuse methods focus on high confidence detection of exact quotation. They can detect one form of direct influence with near certainty but are more limited to paraphrasing and indirect influence. Mullen and Funk's "Spine of American Law"[6] and Cordell and Smith's *Viral*

---

[2] Li, Zehan, Xin Zhang, Yanzhao Zhang, Dingkun Long, Pengjun Xie, and Meishan Zhang. "Towards general text embeddings with multi-stage contrastive learning." *arXiv preprint arXiv:2308.03281* (2023).

[3] Opitz, Juri, Letitia Parcalabescu, and Anette Frank. "AMR similarity metrics from principles." *Transactions of the Association for Computational Linguistics* 8 (2020): 522-538.

[4] Rockmore, Daniel N., Chen Fang, Nicholas J. Foti, Tom Ginsburg, and David C. Krakauer. "The cultural evolution of national constitutions." *Journal of the Association for Information Science and Technology* 69, no. 3 (2018): 483-494.

[5] Barron, Alexander TJ, Jenny Huang, Rebecca L. Spang, and Simon DeDeo. "Individuals, institutions, and innovation in the debates of the French Revolution." *Proceedings of the National Academy of Sciences* 115, no. 18 (2018): 4607-4612.

[6] Funk, Kellen, and Lincoln A. Mullen. "The spine of American law: Digital text analysis and US legal practice." *The American Historical Review* 123, no. 1 (2018): 132-164.

*Texts*[7] both search a large corpus for direct quotations while using a mix of computational methods to remain robust to OCR errors. While direct quotation detection ensures high confidence, it necessarily only captures a very limited range of potential influence, excluding similarities in language use, indirect quotation, and similar claims.

Finally, approaches focused on detecting similarity and changes in word sense (for example, comparing diachronic embeddings of how concepts like 'justice' evolved over time) can capture stylistic and discursive influence. Soni and Klein's[8] study of Abolitionist newspapers uses word2vec word embeddings. Other approaches, such as Vicinanza et. al.[9] use language models such as BERT to measure stability and innovation in word senses. However, these findings can be very difficult to interpret across entire vocabularies and are unable to capture any changes in content or argumentation. The influence they capture is also highly speculative; stylistic changes may reflect wider shifts in language use instead of direct interactions.

My proposed method attempts to synthesize text reuse and word sense embedding methods. By evaluating claims on the sentence level, we can gain a granular understanding of specific ideas, while also remaining open to abstract similarities in meaning or structure. Specialized sentence embeddings language models have demonstrated improved effectiveness in encoding semantic meaning in general evaluation tasks as compared to standard BERT and Word2Vec embeddings.[10] Sentence embeddings have been applied to the task of detecting citation and plagiarism in general academic literature[11] [12] as well as encoding documents specific to disciplinary subfields.[13] I selected GTE vectorization because of the lower computational demands of the GTE-small model and its superiority in evaluation metrics to other sentence embedding methods.

---

[7] https://manifold.umn.edu/projects/going-the-rounds
[8] Soni, Sandeep, Lauren Klein, and Jacob Eisenstein. "Abolitionist networks: Modeling language change in nineteenth-century activist newspapers." *arXiv preprint arXiv:2103.07538* (2021).
[9] Vicinanza, Paul, Amir Goldberg, and Sameer B. Srivastava. "A deep-learning model of prescient ideas demonstrates that they emerge from the periphery." *PNAS nexus* 2, no. 1 (2023): pgac275.
[10] Reimers, Nils, and Iryna Gurevych. "Sentence-bert: Sentence embeddings using siamese bert-networks." *arXiv preprint arXiv:1908.10084* (2019).
[11] Alvi, Faisal, Mark Stevenson, and Paul Clough. "Paraphrase type identification for plagiarism detection using contexts and word embeddings." *International Journal of Educational Technology in Higher Education* 18, no. 1 (2021): 42.
[12] Lagopoulos, Athanasios, and Grigorios Tsoumakas. "Self-citation Analysis using Sentence Embeddings." *arXiv preprint arXiv:2105.05527* (2021).
[13] Chen, Qingyu, Yifan Peng, and Zhiyong Lu. "BioSentVec: creating sentence embeddings for biomedical texts." In *2019 IEEE International Conference on Healthcare Informatics (ICHI)*, pp. 1-5. IEEE, 2019.

**Data:**

To evaluate this method, I constructed a dataset based around authors active in 19th century academic societies in the British Empire. I curated a list of journals based on secondary readings and prior knowledge.[14] [15] This is not meant to capture comprehensively all academic publications in the 19th century, but rather to gather a representative cross section of the most active members of this community. Below is a list of the journals scraped:

- General (5000 matches):
    - Royal Society (including colonial branches)
    - Royal Institution
    - Cambridge Philosophical Society
- Chemical (337 matches):
    - (London) Chemical Society
- Medical (77 matches):
    - (London) Medical and Chirurgical Society
- "Natural History" (2657 matches):
    - Linnean Society
    - Zoological Society
    - Entomological Society
    - Geological Society
- Geography (592 matches):
    - Geographical society
- Political/social scientific (1333 matches)
    - The Economist
    - Westminster Review
    - Edinburgh Review
- Orientalist (1570 matches):
    - Asiatic Society: Royal, Calcutta, American

I grouped these societies into proto-disciplines such as biology, geology, chemistry, and politics/social science. I constructed a supplementary dataset of books by Darwin's correspondents using letters from the Darwin Project[16]. Author names were extracted from downloaded proceedings using Spacy's NER utility. ~400,000 books by the ~1.000,000 identified potential authors were downloaded as digitized texts from the Internet Archive and Project Gutenberg. Metadata for the corpus is available at this link.[17] I also used the Project Gutenberg editions of Darwin's *Origin of Species* and *Descent of Man* and Herbert Spencer's *Principles of Sociology* and *Principles of Biology* for a comparative sample.

---

[14] Pal, Eszter. "Scientific societies in Victorian England." *Review of Sociology* 20 (2014): 85-111.
[15] Barton, Ruth. "'An Influential Set of Chaps': The X-Club and Royal Society Politics 1864–85." *The British journal for the history of science* 23, no. 1 (1990): 53-81.
[16] https://www.darwinproject.ac.uk/
[17] https://uofi.box.com/s/i4024nqx70zjylgqd4nmackqvqsuusvj

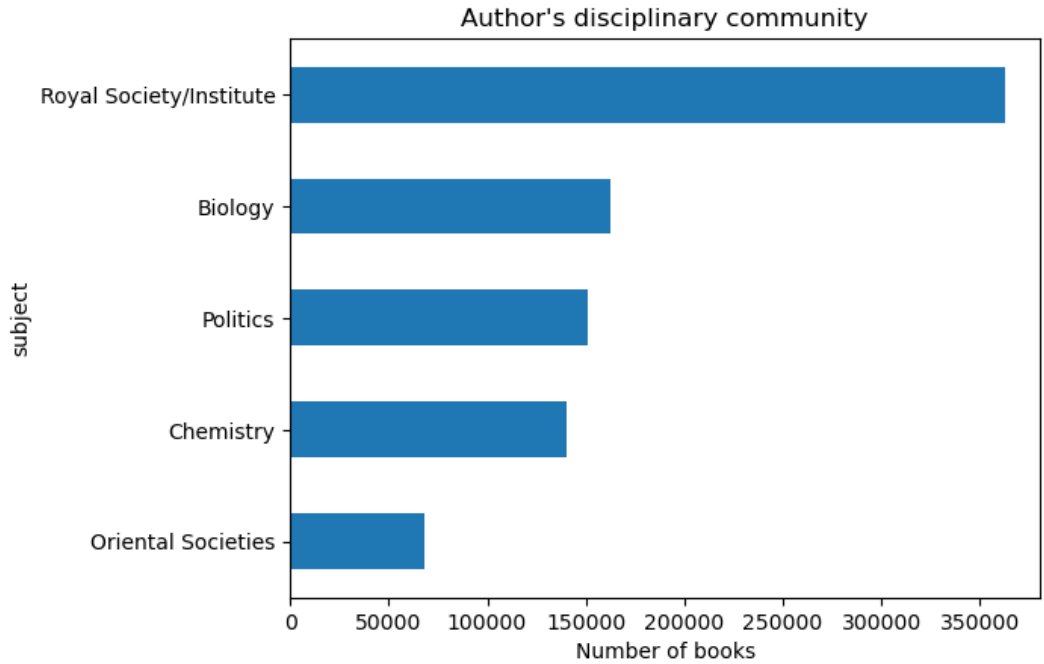

*Fig 1. number of books identified for each disciplinary community. There is a high degree of overlap between these categories.*

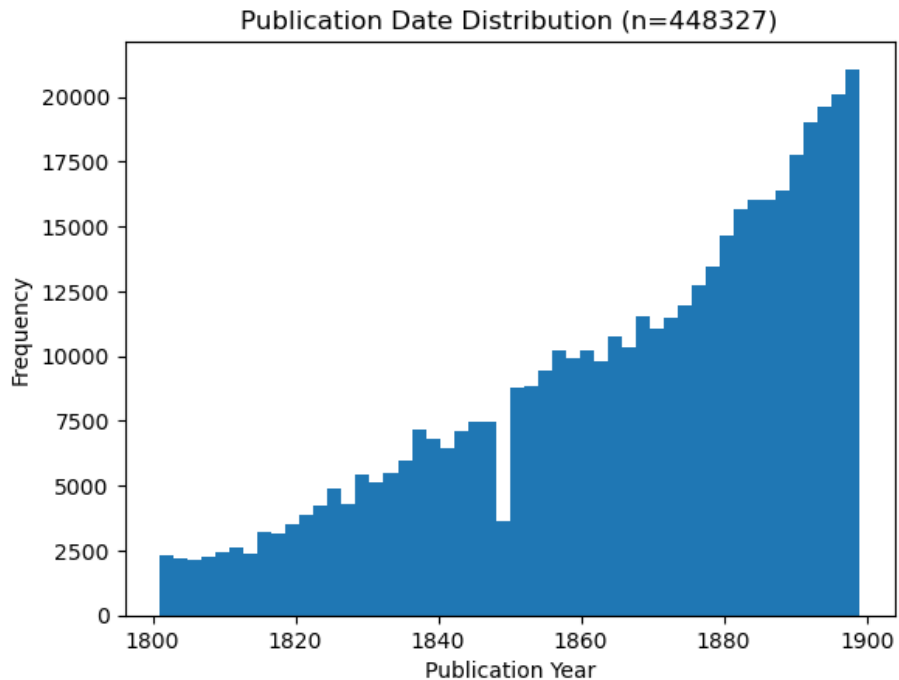

*Fig 2. distribution of corpus books by publication year. The linearly increasing trend over time is a product of the Internet Archive's collection and digitization bias as well as the name selection methodology.*

**Method:**

I performed sentence tokenization per book using NLTK. Overly short documents (<1000 characters) and sentences (<45 words) were removed because short documents tended to be mislabeled documents or contents not amenable to OCR. Short sentences tended to not contain enough information for a coherent argument. Each sentence was converted into an embedding using an ensemble of language model methods.

Initially embeddings were generated with General Text Embeddings (GTE), a BERT based approach fine-tuned with internet text and specific entailment tasks to capture semantic meaning. GTE embeddings were generated for each sentence using the GTE-small model implemented in the sentence-transformers Python package[18]. GTE-small was selected due to memory and computational power constraints. No additional fine tuning or hyperparameter changes were performed. Using these vectors, I used FAISS[19] to create rapidly searchable cosine indices for every sentence in the corpus.

For the books of interest (*Origin of Species, Descent of Man, Principles of Sociology,* and two randomly sampled books, *Form of the Horse* and *Echoes from the Backwoods,* published in the same year as *Origin)* the tokenization and vectorization process were repeated. Each sentence of the books of interest were queried against every other sentence in the corpus based on cosine similarity to discover potential matches. A more detailed diagram of dataset curation and methods is included in supp. fig. 1. For further analysis, I used thresholds of >0.85 cosine similarity (speculative and low confidence), >0.90 cosine similarity (indirect/medium influence) and >0.95 (high confidence and direct quotation).

After books with a high number (>5) of semantic matching sentences with *Origin of Species* were discovered, evaluations of other similarity methods were conducted. Structural comparisons using abstract meaning representation (AMR) graphs provide potential to capture similarities in argument structure and metaphor. I used the python amr-lib[20] package to generate graphs for each sentence, which can be compared for similarities in graph structure. Currently, the majority of the corpus has not been analyzed using this method, as both the AMR graph generation and graph matching are prohibitively computationally intensive. However, there is significant space for future exploration in this direction.

---

[18] https://www.sbert.net/
[19] https://github.com/facebookresearch/faiss
[20] https://github.com/bjascob/amrlib

This ensemble approach will hopefully represent both the specific claim advanced and rhetorical and argumentative strategies supporting the claim. Code for the entire pipeline is available in this GitHub repository.[21]

**Findings:**

Because annotated data does not exist for the very messy corpus of scanned 19th century books, I conducted evaluation against historical ground truth.

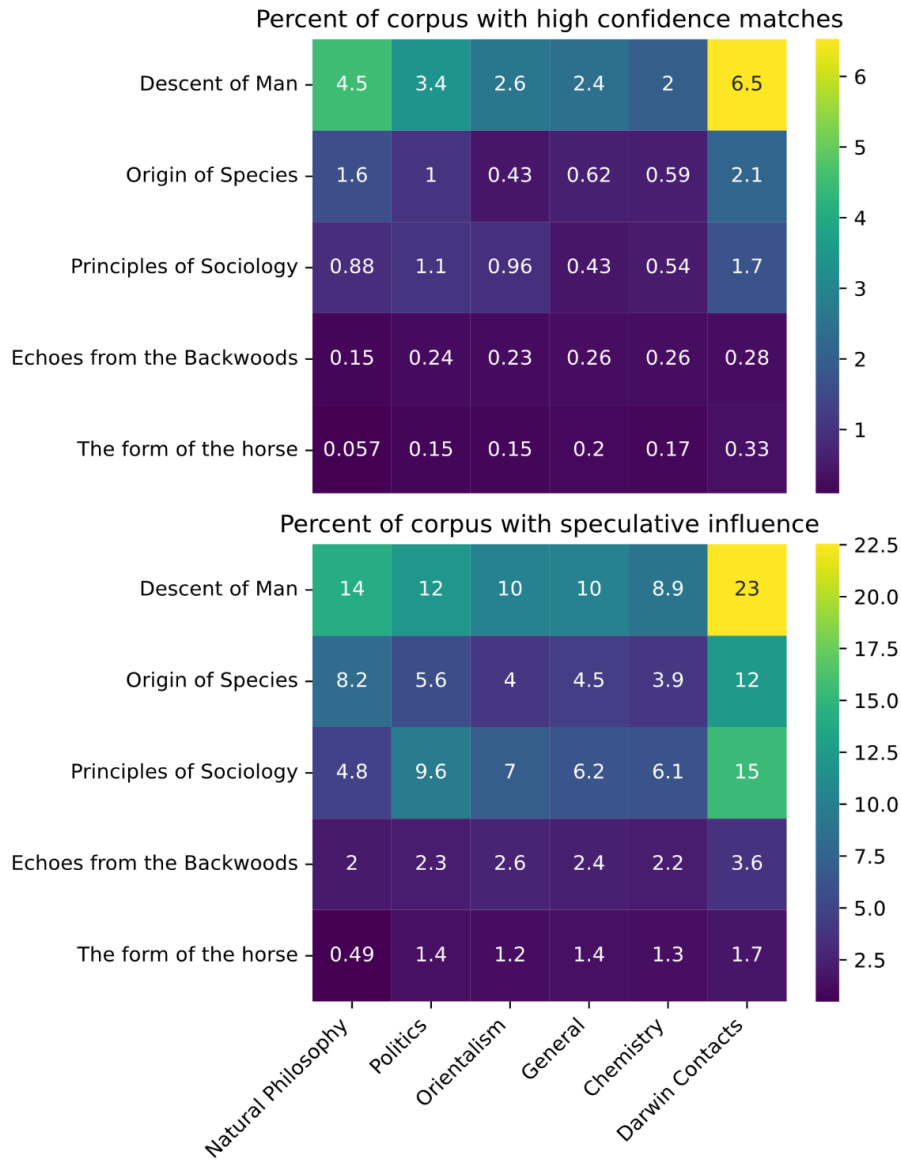

Fig. 2: percent of books post-publication with detected influence. The last two books are randomly sampled and included as a baseline comparison.

---
[21] https://github.com/lucianli123/darwin-novelty

The rightmost column of Figure 2 shows the overrepresentation of Darwinian influence in books by Darwin's correspondents (people with documented interactions with Darwin). As a further confirmation, we can see more influence from Darwin's books in Biology and Geology than Chemistry or Political Theory. Even when the confidence threshold is lowered and more speculative matches are allowed, the same patterns persist.

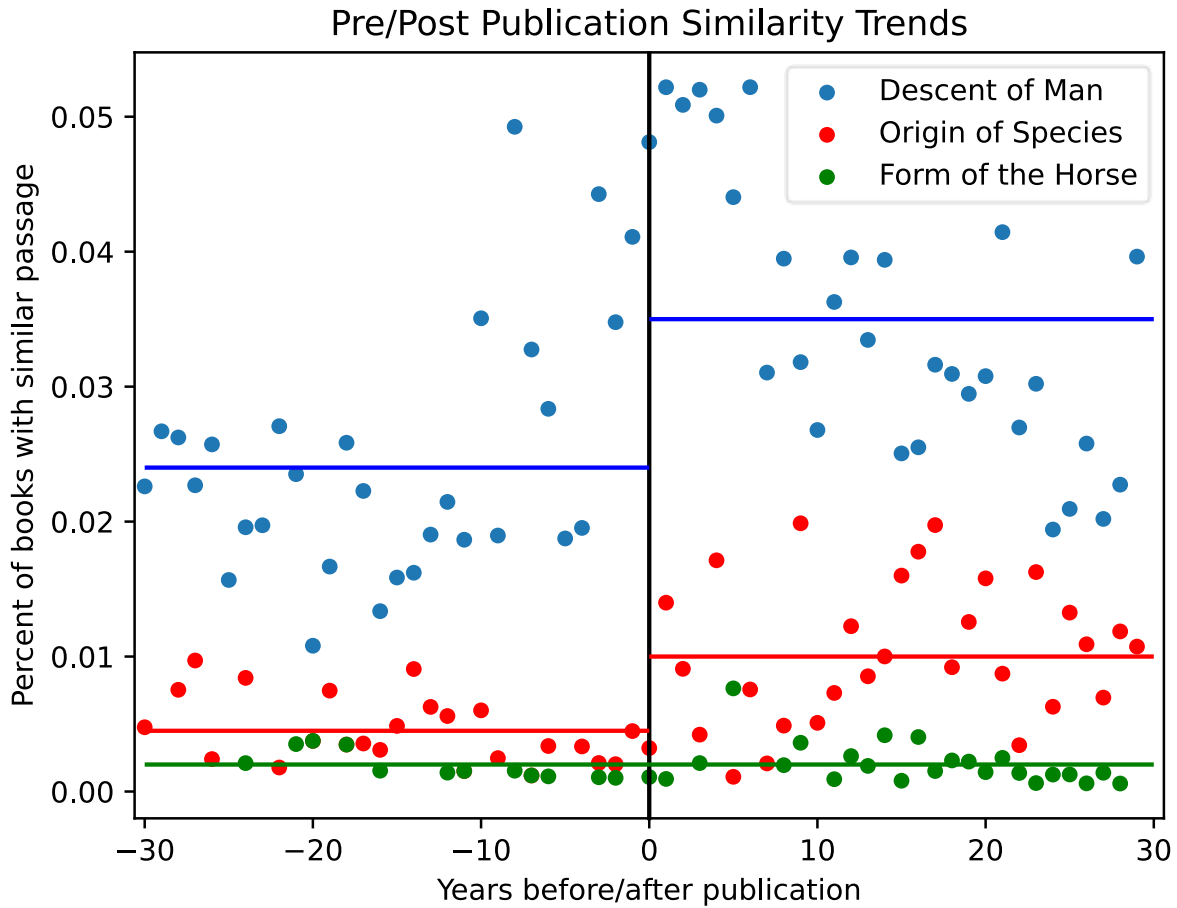

*Fig 3. Similarity with pre and post publication books. Green book included as baseline*

When we plot the influence over time, we get further confirmation of the method's effectiveness. Each point represents the similarity of books published each year and the colored lines represented the average similarity of all pre and post publication books. In red, the *Origin of Species* (1859) draws from a handful of primarily geological and biological sources pre-publication, but radically shifts the discourse. In blue, the *Descent of Man (*1871) engages more with discourses across a diverse range of disciplines as well as the evolutionary ideas already introduced in the *Origin*. However, it likewise radically shifts the discourse across scholarly circles.

| Sentence 1 | Sentence 2 | Sentence Embedding Cosine Distance |
|---|---|---|
| Would it be believed, that the larvae of an insect, or fly, no larger than a grain of rice, destroy some thousand acres of pine-trees, many of them from two to three feet in diameter, and a hundred and fifty in height ? | ** Would it be believed,*^ says Wilson, the ornitholog-ist, '* that the larvs of an insect, or fly, no larger tlia^n a grain of rice, should, destroy some thousand ncres of pine trees, many of uiem two or three feet in diameter, and one himdred and fifty feet high. | 0.97 |
| I have called this principle, by which each slight variation, if useful, is preserved, by the term natural selection, in order to mark its relation to man's power of selection. | The expression "natural selection" was chosen as serving to indicate some parallelism with artificial selection--the selection exercised by breeders. | 0.92 |
| I have so much to do in unraveling certain human lots, and seeing how they were woven and interwoven, that all the light I can command must be concentrated on this particular web, and not dispersed over that tempting range of relevancies called the universe. | We shall never disentangle the inextricable web of affinities between the members of any one class; but when we have a distinct object in view, and do not look to some unknown plan of creation, we may hope to make sure but slow progress. | 0.85 |

*Table 1: examples of sentences detected at various similarity thresholds by the proposed method. In the first example, we can see that the method detects direct quotation at high confidence while remaining robust to OCR errors and minor structural changes. The second example shows the ability of the method to identify cases of paraphrase with very limited shared word use. Lastly, we see the ability of the method to capture speculative matches across disciplines by returning to our epigraph quotations.*

**Conclusion:**

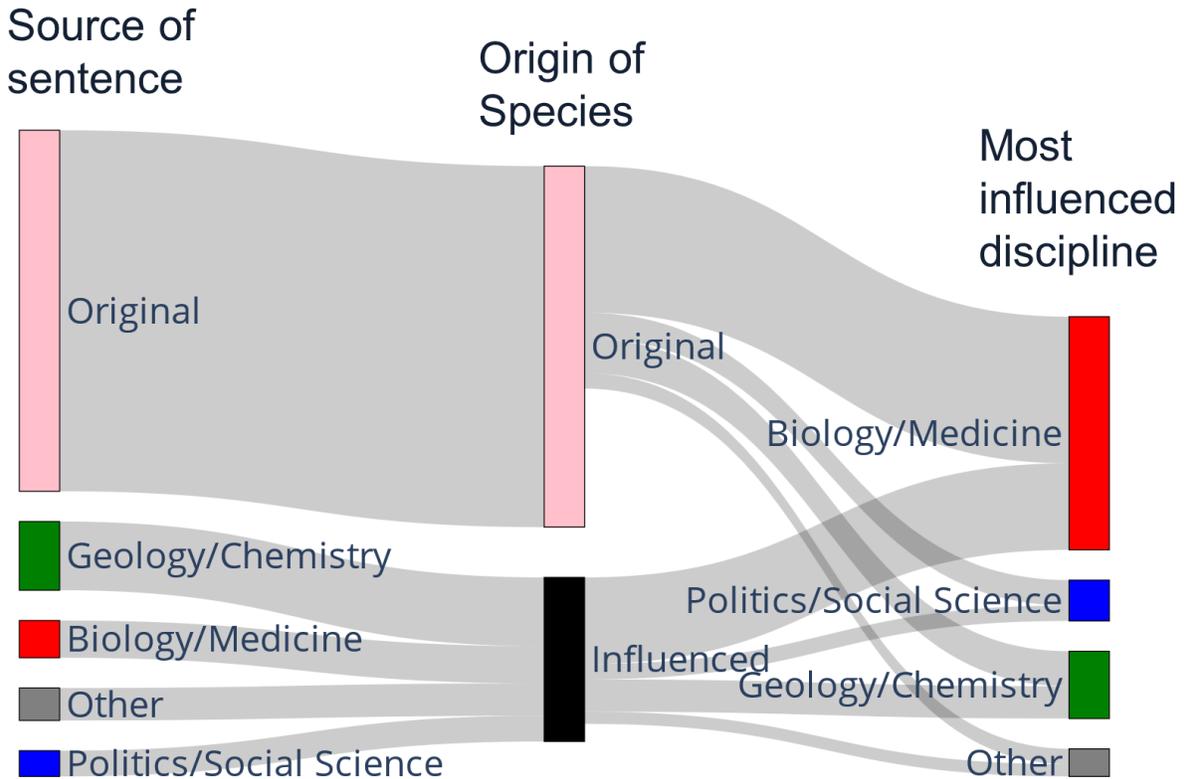

*Fig 4. Alluvial diagram of where claims in* Origin of Species *originated and gained traction. Starting from* OoS *in the center, we can consider previously published books to see where each claim originated. We can also consider the afterlives of the statements; which disciplines certain claims resonated strongly with and how they were used.*

      This method allows for a hypertextual exploration of any given text. Imagine an edition of the *Origin of Species* where a reader can click each sentence and receive information on where that argument appeared pre-publication. They would be able to observe the heavy influences from geology, as well as Darwin's own original observations based on his travels. A reader would then be able to look forward and see which fields each statement resonated with and the context for how they read and interpreted sections differently – in our Darwinian case, similarities and differences in ways eugenicists read the *Origin* compared to botanists. Now imagine this on a larger scale: instead of arguments from the *Origin,* all arguments in the corpus. Would we be able to find common features of ideas which gained wider traction or leapt across disciplinary boundaries?

Traditional narratives of discovery and invention valorize the contributions of individual geniuses - almost exclusively wealthy men from metropolitan societies. While historians of science have challenged this paradigm, dependence on personal papers and close reading of related works limits the potential scale and representativeness of these efforts. Even Beer's incisive work ultimately limits itself to Anglo-American literature and canonical authors. Responsible use of potentially destabilizing new AI technologies, keeping in mind their gaps and exclusions, can radically reshape our view of genealogies of ideas and influence and suggest previously unimagined possibilities for further exploration.

This mode of analysis has the potential to uncover connections between the work of hundreds of thousands of authors, among them women explorers and scientists, interlocutors from colonized peoples, and simply those whose ideas and contributions have been forgotten in the present. These ideas are as much part of the patchwork of intellectual life in the 19th century as those of Darwin or Herbert Spencer or Charles Lyell. Taking a wider view has the potential to reinvent the history of science.

ideas demonstrates that they emerge from the periphery." *PNAS nexus* 2, no. 1 (2023): pgac275.

Supplemental Figure 1:

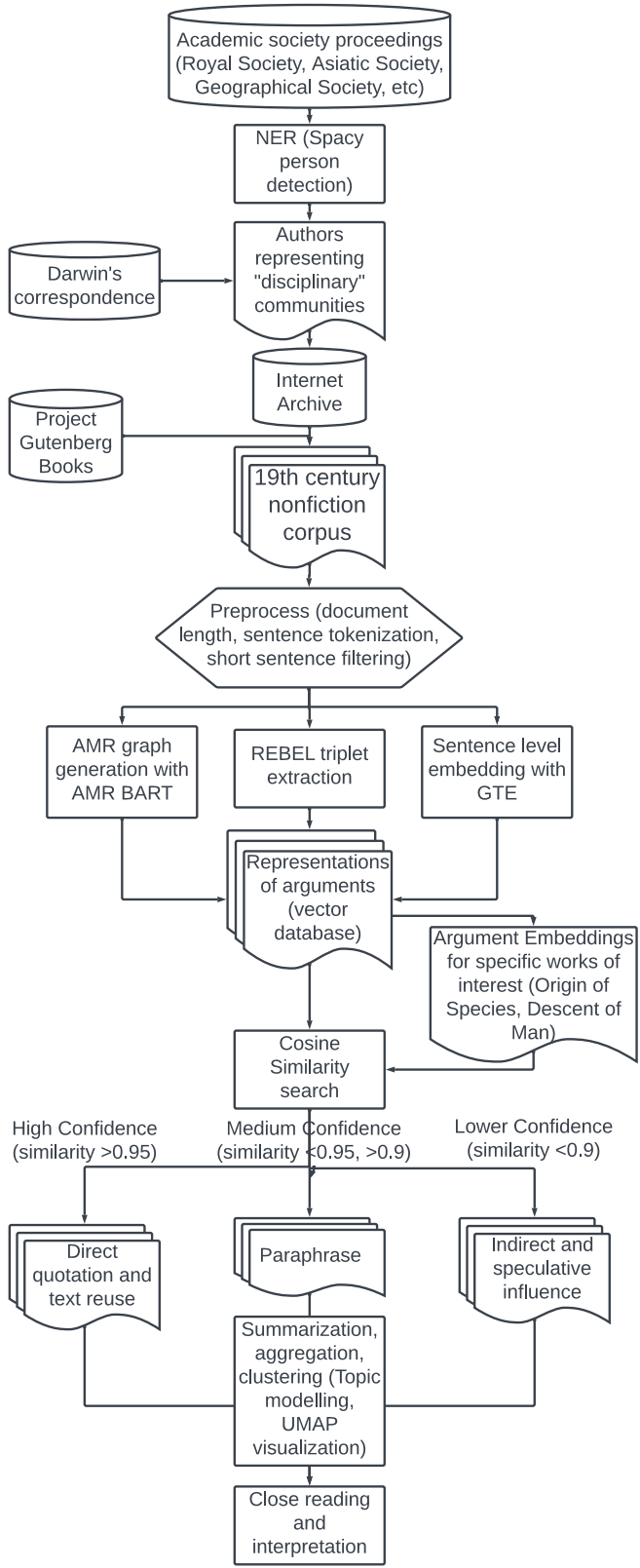